\documentclass[conference]{IEEEtran}
\IEEEoverridecommandlockouts
\usepackage{cite}
\usepackage{amsmath,amssymb,amsfonts}
\usepackage{algorithmic}
\usepackage{graphicx}
\usepackage{textcomp}
\usepackage{xcolor}
\usepackage{array}
\usepackage{colortbl}
\usepackage{diagbox}
\usepackage{footnote}
\usepackage{hyperref}

\def\BibTeX{{\rm B\kern-.05em{\sc i\kern-.025em b}\kern-.08em
    T\kern-.1667em\lower.7ex\hbox{E}\kern-.125emX}}

\makeatletter
\newcommand{\thickhline}{%
	\noalign {\ifnum 0=`}\fi \hrule height 2pt
	\futurelet \reserved@a \@xhline
}
\newcolumntype{"}{@{\hskip\tabcolsep\vrule width 2pt\hskip\tabcolsep}}
\makeatother

\begin{document}

\title{Mutual Modality Learning for Video Action Classification}

\author{
\IEEEauthorblockN{Stepan Komkov, Maksim Dzabraev, Aleksandr Petiushko}
\IEEEauthorblockA{
Lomonosov Moscow State University\\
Huawei Moscow Research Center\\
stepan.komkov@intsys.msu.ru, dzabraev.maksim@intsys.msu.ru, petyushko.alexander1@huawei.com}
}

\maketitle

\begin{abstract}
	
The construction of models for video action classification progresses rapidly. However, the performance of those models can still be easily improved by ensembling with the same models trained on different modalities (e.g.~Optical flow). Unfortunately, it is computationally expensive to use several modalities during inference. Recent works examine the ways to integrate advantages of multi-modality into a single RGB-model. Yet, there is still a room for improvement. In this paper, we explore the various methods to embed the ensemble power into a single model. We show that proper initialization, as well as mutual modality learning, enhances single-modality models. As a result, we achieve state-of-the-art results in the Something-Something-v2 benchmark.
\end{abstract}


\section{Introduction}

Video Recognition has progressed a lot during the last several years. Datasets have enlarged from thousands of clips \cite{kuehne2011hmdb, soomro2012ucf101} to hundreds of thousands \cite{kay2017kinetics,carreira2019short,goyal2017something} and even to hundreds of millions \cite{miech2019howto100m}. Neural network-based approaches for video processing evolved from simple 3D-convolutions \cite{tran2015learning} to Parvo- and Magnocellular counterparts emulation \cite{feichtenhofer2019slowfast} and absorbed developments of classical Image Recognition \cite{carreira2017quo,lin2019tsm}. 

Nevertheless, classical results in the domain of Video Processing are still useful: Optical Flow estimation for a video sequence can significantly improve the quality of video recognition \cite{simonyan2014two}. However, the common ways to estimate Optical Flow require an amount of calculation that is comparable to the whole further neural network inference. That is why a number of works are devoted to the implicit Optical Flow estimation during the RGB-based neural network inference \cite{fan2018end, Crasto_2019_CVPR,piergiovanni2019representation,Stroud_2020_WACV}.

In our work, we target not only the improvement of RGB-based models but also simultaneous improvement of different single-modality models. To this end, we utilize Mutual Learning \cite{zhang2018deep} that enables us to share the knowledge between single-modality models in both directions. We combine it with proper initialization that develops the performance of trained models.

We show that our approach not only improves each single-modality model but also boosts RGB-based models better than existing methods. Additionally, we examine how to use Mutual Learning to achieve the best results of multi-modality ensemble. Thus, we achieve state-of-the-art (SOTA) results among the ones reported previously in the Something-Something-v2 benchmark \cite{goyal2017something}.

Our code is available as a fork from the code presented in \cite{lin2019tsm} (\url{https://github.com/papermsucode/mutual-modality-learning}).

\section{Related Works}

First, we briefly describe the most common approaches for video action classification from the historical perspective. They can be divided into two groups: models with 3d-convolutions and models with 2d-convolutions.

Second, we describe the methods that improve single-modality models using other modalities and highlight the differences from our work.

\subsection{3D-approaches}

A video sequence is a 4d-tensor with the following parameters: height of frames, width of frames, number of frames, and number of channels per frame (3 in case of RGB input). Therefore, we can process it using Convolutional Neural Networks (CNNs) where 3d-convolutions are applied instead of 2d convolutions (with the new temporal dimension). Tran \textit{et al.}~are the first to propose 3d convolutional networks based on this idea \cite{tran2015learning}. Thereby, they achieved the SOTA results in a number of tasks.

Although a video model has to obtain temporal and motion information from the sequence of frames, it still needs to recognize the spatial information contained in each frame. Carreira and Zisserman propose to inflate the trained weights of the Image Recognition network and to use them as initialization for 3d-CNN \cite{carreira2017quo}. Nowadays, this is a common approach for video model initialization.

Wang \textit{et al.}~implement an attention mechanism that helps to find dependencies between far positions on different frames. That is meaningful for fast-moving objects or quick movements of the camera \cite{wang2018non}.

The disadvantage of 3d-CNNs is that they require to work with much more parameters in comparison to their 2d analogs. To address this problem, the first convolutions can be replaced by the per-frame 2d-convolutions (top-heavy models) since those convolutions are mostly responsible for the evaluation of spatial features \cite{xie2018rethinking,zolfaghari2018eco}. Also, 3d-convolutions can be decomposed as 2d spatial-convolutions plus 1d temporal-convolutions. This kind of decomposition reduces the number of parameters and operations and increases non-linearity at the same time \cite{xie2018rethinking,tran2018closer}.

Feichtenhofer \textit{et al.}~present a SlowFast Network architecture that emulates Parvo- and Magnocellular counterparts by sampling video frames with two different framerates and by feeding them to two branches with different computational power \cite{feichtenhofer2019slowfast}. Thus, the lightweight Fast pathway captures motion and temporal dynamics while the Slow pathway captures the spatial semantics. This approach achieved the SOTA results on the Kinetics-400 Action Classification dataset \cite{kay2017kinetics} among models without additional data.

The Temporal Pyramid Networks (TPN) of Yang \textit{et al.}~can be viewed as an extension of SlowFast networks \cite{yang2020temporal}. A thinned out frames sequence flows to the different branches from intermediate layers instead of entering from the input. This approach is an add-on to existing architectures and can be implemented for 3d-CNNs and 2d-CNNs.

\subsection{2D-approaches}

The early CNN-based models for video with 2d-convolutions consist of two streams. The first stream called Spatial takes RGB frames as an input. The second stream called Temporal takes a stack of consecutive Optical Flow estimations \cite{simonyan2014two,wang2016temporal}. The final prediction is an average of the predictions of both streams. Note that the authors use pretrained weights from Image Recognition models for the temporal stream as well as for the spatial stream. 

Nowadays, the idea of features sharing between frames is used to simulate a 3d-inference using 2d-convolutions. The pioneering work in this scope is Temporal Shift Modules network (TSM) by Lin \textit{et al.}~that applies ordinary 2d-ResBlocks \cite{he2016deep} to each input frame \cite{lin2019tsm}. The single difference is that TSM replaces a one-eighth of channels with the same channels from the previous frame and another one-eighth of channels with the same channels from the future frame before each first convolution of the ResBlock. Thereby, the authors achieved the SOTA results on the Something-Something-v2 dataset \cite{goyal2017something} and provided a powerful and efficient baseline for the future research in this area.

Based on the idea of feature sharing, Shao \textit{et al.}~present Temporal Interlacing Network \cite{shao2020temporal}. The authors add extra lightweight blocks that decide on distances and weights of channels sending within each ResBlock. This approach is used instead of a fixed replacement of channels.

\subsection{Optical Flow distillation}

Despite all the aforementioned progress, most of works can be improved by averaging of their predictions with the predictions of the same network trained on the Optical Flow modality \cite{carreira2017quo,xie2018rethinking,tran2018closer,zolfaghari2018eco,wang2016temporal,lin2019tsm}.

Since the Optical Flow calculation is a time-consuming operation, a number of works is devoted to incorporation of the motion-estimation blocks inside the CNN architecture \cite{fan2018end,piergiovanni2019representation,jiang2019stm}. However, knowledge distillation from the Optical Flow modality to any RGB single-modality network seems to be of more interest.

Three basic works that should be mentioned are Knowledge Distillation (KD) \cite{hinton2015distilling}, Mutual Learning (ML) \cite{zhang2018deep}, and Born-Again Networks (BAN) \cite{furlanello2018born}.

The first proposes to use soft-predictions of the model called Teacher network to train the smaller model called Student network. It turns out that this technique is helpful for video action classification task not as a neural network compression method but as a transfering of modality knowledge. Zhang \textit{et al.}~use KD to train a two-stream network with Motion Vector as the second modality \cite{zhang2016real}. Stroud \textit{et al.}~confirm by constructing Distilled 3D Networks (D3D) \cite{Stroud_2020_WACV} that KD from the Optical Flow stream improves the quality of the RGB stream. In addition, the authors of D3D show that KD teaches implicit Optical Flow calculation inside the RGB stream. Motion-Augmented RGB Stream (MARS) of Crasto \textit{et al.}~distills the knowledge not from the prediction of the Optical Flow stream but from its feature maps before the global averaging operation \cite{Crasto_2019_CVPR}.

In contrast to the mentioned works, we utilize the idea of ML to train jointly several single-modality networks and improve the quality of each of them. Motivated by BAN, we show that the relaunch of training procedure can further boost the performance of models. Additionally, we show that proper initialization improves our results as well as results for MARS and D3D works.

Note that we target on the single-modality model quality. The improvement of the average predictions of several streams is a different branch of research. An example of an approach that addresses this problem is Gradient-Blending \cite{wang2019makes}. Nevertheless, we examine the ability of ML to improve the average prediction the multi-modality ensemble. It turns out that proposed initialization with relaunches of single-modality ML provides the best result for the ensemble.

\begin{figure*}[!t]
	\centering
	\includegraphics[width=\linewidth]{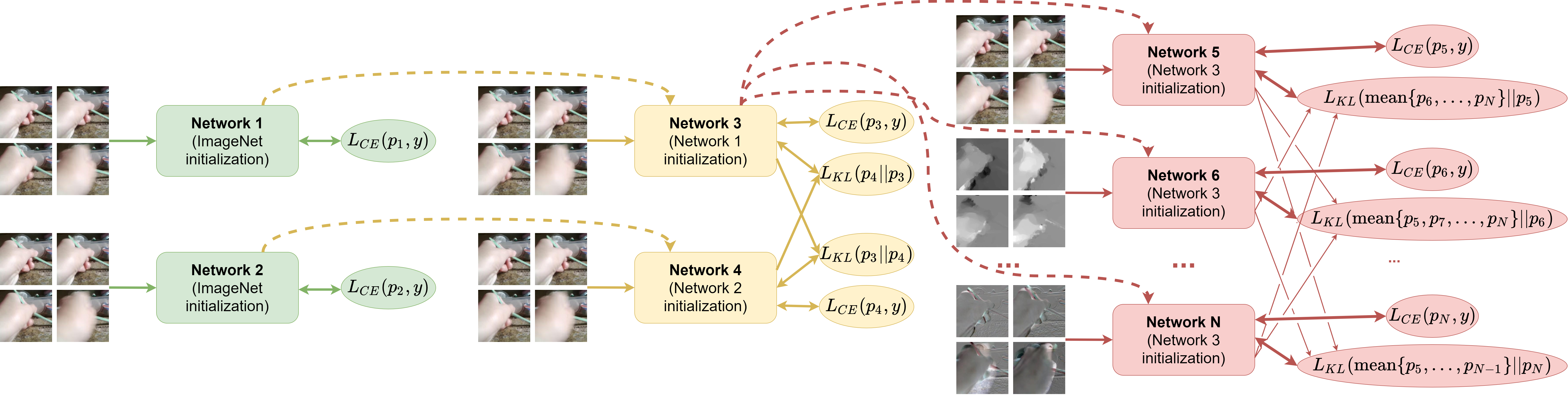}
	\caption{Best viewed in color. Solid arrows denote flows of data. Dashed arrows denote weights transferring for initialization.
		Green part: first, we train two networks with RGB input initialized by ImageNet weights using cross-entropy loss.
		Yellow part: next, we use weights from the first step as initialization for two networks with RGB input that are trained jointly using Mutual Learning.
		Red part: finally, we apply Mutual Modality Learning to obtain the best single-modality model for each modality. We use weights of the network from the second step as initialization for each model in the third part.}
	\label{fig:solo}
\end{figure*}

\section{Proposed solution}

The proposition of the best single-modality model training pipeline is depicted in Figure~\ref{fig:solo}. The pipeline for the best ensemble training is described in section \ref{sec:ensemble}.

The pipeline consists of three parts: initialization preparation, ML implantation and Mutual Modality Learning (MML).

The importance of each part is confirmed in section \ref{sec:ablation}.

\subsection{Initialization preparation}

The standard starting point for the Video Action Classification models training is an ImageNet \cite{deng2009imagenet} pretrained model. Inflating of 2d-convolutions proposed in \cite{carreira2017quo} makes that possible for both 3d-models and 2d-models.

If we use the input modality different from RGB then we have to change the shape of the first convolution from $(C,3,K,K)$ to $(C,N,K,K)$. Here, $C$ is a number of output channels of the first convolution, $K$ is a kernel size and $N$ is a number of channels of the new input. The pseudocode for the weights of the new convolution is as follows:

\begin{verbatim}
for i in 1:N do
    W_new[:,i] = (W[:,1]+W[:,2]+W[:,3])/3
end
\end{verbatim}

In the proposed pipeline, we use ImageNet initialization only for the first step. The next two steps use weights from the previous step (with a change in the first convolution shape if it is needed).

\subsection{Mutual Learning implantation}

ML is a technique of training two models together in a way that they help each other to reach better convergence. To achieve that, we modify the loss functions of the networks as follows:
\begin{equation}
	L_1 = L_{CE}(p_1,y)+L_{KL}(p_2||p_1),
\end{equation}
\begin{equation}
	L_2 = L_{CE}(p_2,y)+L_{KL}(p_1||p_2).
\end{equation}
Here, $L_i$ is a loss of the $i$-th network, $p_i$ is a vector of the predicted class probabilities by the $i$-th network, $y$ is a ground-true class label, $L_{CE}$ is a cross-entropy loss and $L_{KL}$ is the Kullback Leibler (KL) Divergence loss given by the formula
\begin{equation}
	L_{KL}(p_i||p_j) = \sum_{n=1}^{N}p_i^n\cdot\log\frac{p_i^n}{p_j^n}.
\end{equation}
In this formula $p_i^n$ stands for a probability for the $n$-th class predicted by the $i$-th model. Thus, models teach each other using dependencies that they found during training and thereby improve their performance.

If there are more than two models involved in ML then the loss function is
\begin{equation}
	L_i = L_{CE}(p_i,y)+L_{KL}\left(\frac{\sum_{j\neq i} p_j}{M-1}||\,p_i\right)
\end{equation}
where $M$ is a number of models. 

\begin{table*}[!t]
	\small
	\begin{minipage}{0.48\textwidth}
		\centering
		\caption{One model training}
		\label{tab:solo}
		\centering
		\begin{tabular}{|m{0.25\columnwidth}|c|m{0.25\columnwidth}|c|} 
			\hline \it Model & \it Top-1 / Top-5 & \it Model & \it Top-1 / Top-5 \\ \hline
			\it RGB from ImageNet & \bf 58.10 / 84.61 & \it RGB from Flow & 57.53 / 84.42 \\ \thickhline
			\it Flow from ImageNet & 52.32 / 81.84 & \it Flow from RGB & \bf 55.19 / 84.14\\ \thickhline
			\it Diff from ImageNet & 58.74 / 84.39 & \it Diff from RGB & \bf 58.98 / 86.33 \\ \hline
		\end{tabular}
	\end{minipage}\hfill
	\begin{minipage}{0.48\textwidth}
		\centering
		\caption{MARS and D3D training of TSM}
		\label{tab:mars}
		\centering
		\begin{tabular}{|m{0.23\columnwidth}|m{0.23\columnwidth}|m{0.23\columnwidth}"m{0.23\columnwidth}|} 
			\hline \it Model & \it Teacher modality & \it MARS training Top-1 / Top-5 & \it D3D training Top-1 / Top-5 \\ \hline
			\it RGB from ImageNet & \it Flow & 57.56 / 84.39 & 58.99 / 85.18 \\ \hline
			\it RGB from RGB & \it Flow & \bf 59.11 / 85.24 & \bf 59.95 / 85.86 \\ \thickhline
			\it Flow from ImageNet & \it RGB & 57.46 / 85.01 & 55.04 / 83.36 \\ \hline
			\it Flow from RGB & \it RGB & \bf 58.23 / 85.37 & \bf 56.41 / 83.98 \\ \hline
		\end{tabular}
	\end{minipage}
\end{table*}

\begin{table*}[!t]
	\begin{minipage}{0.48\textwidth}
		\centering
		\caption{RGB results of MML}
		\label{tab:rgbml}
		\centering
		\begin{tabular}{|m{0.23\columnwidth}|m{0.23\columnwidth}"m{0.23\columnwidth}"m{0.23\columnwidth}|} 
			\hline \it RGB results & \it Flow from ImageNet & \it Flow from Flow & \it Flow from RGB \\ \hline
			\it RGB from ImageNet & 56.25 / 84.07 & 60.02 / 86.08 & 58.70 / 85.18 \\ \hline
			\it RGB from RGB & \bf 60.80 / 86.47 & \bf 60.94 / 86.67 & \bf 60.82 / 86.75 \\ \hline
			\it RGB from Flow & 58.37 / 84.82 & 58.56 / 85.28 & 58.62 / 85.36 \\ \hline
		\end{tabular}
	\end{minipage}\hfill
	\begin{minipage}{0.48\textwidth}
		\centering
		\caption{Flow results of MML}
		\label{tab:flowml}
		\centering
		\begin{tabular}{|m{0.23\columnwidth}|m{0.23\columnwidth}|m{0.23\columnwidth}|m{0.23\columnwidth}|} 
			\hline \it Flow results & \it Flow from ImageNet & \it Flow from Flow & \it Flow from RGB \\ \hline
			\it RGB from ImageNet & 54.94 / 83.82 & 57.06 / 84,87 & \bf 57.84 / 85.15 \\ \thickhline
			\it RGB from RGB & 54.76 / 83.61 & 56.74 / 84.78 & \bf 57.95 / 85.44 \\ \thickhline
			\it RGB from Flow & 55.85 / 84.33 & 56.86 / 84.80 & \bf 57.79 / 85.34 \\ \hline
		\end{tabular}
	\end{minipage}
\end{table*}

\subsection{Mutual Modality Learning}

In the original ML, both models use the same modality as an input. We propose to use different modalities of the video obtained from the same frames as inputs for different models. Thus, we share the knowledge obtained from one modality to other modalities.

Note that we need two consecutive frames to calculate the Optical Flow. Thus, if there are $N$ RGB frames in total then there are only $N-1$ Optical Flow frames in total.

So, suppose that the model requires $T$ input-frames for the prediction and we have two representations of the video by different modalities: one representation with $n$ frames and another with $N$ frames ($N>n$).

For this and similar cases in our work, we first sample frames with numbers $(i_1,\ldots,i_T)$ for the modality with the least number of frames, and then we use frames with numbers $(i_1+\xi,\ldots,i_T+\xi)$ for the modality with the biggest number of frames. Here $\xi\sim\text{unif}\{0,\ldots,N-n\}$.

\section{Ablation studies} \label{sec:ablation}

There are several conclusions that we make:
\begin{itemize}
\item Initialization with the RGB model trained on the same video dataset significantly enhances the performance for various modalities and training scenarios (not only ML but MARS and D3D also).
\item MML performs better than MARS or D3D approaches.
\item Two iterations of ML are better than one and there is no need for more.
\item MML performs better than ML as a final step.
\item The behavior described above preserves when we use modalities different from the Optical Flow.
\end{itemize}

\subsection{Experiments setup}

For the ablation studies, we use several models and benchmarks: TSM \cite{shao2020temporal} on Something-Something-v2 \cite{goyal2017something} with the code provided by the authors (the main setup, we use it unless otherwise specified) and I3D \cite{carreira2017quo} on Charades \cite{Sigurdsson2016HollywoodIH} with the code provided in \url{https://github.com/facebookresearch/SlowFast}.

We obtain the Optical Flow using TV-L1 algorithm \cite{zach2007duality} and combine 5 consecutive evaluations of the Optical Flow by the x- and y- axes as one input-frame.

For the RGBDiff modality, we take 6 consecutive RGB frames to obtain 5 consecutive differences between them. Obtained differences are concatenated and considered as one input-frame.

\subsubsection{TSM on Something-Something-v2}

We use the standard setup for the TSM+ResNet-50 \cite{he2016deep} training proposed by the authors with batch size 64, ImageNet pretrain, 0.025 initial learning rate. The only difference is the frames sampling strategy. Instead of using one sampling strategy, we use both uniform sampling and dense sampling. The first one works as follows: we split video into $T$ equal parts and take a random frame from each of them. Dense sampling requires taking each $\tau$-th frame starting from a random position. We apply each of the two sampling strategies with $50\%$ probability. See Appendix~\ref{app:a} as an explanation for this strategy.

We use single uniform sampling with one spatial 224x224 center crop during testing for the ablation studies. That is why the baseline result is worse than the same in \cite{lin2019tsm} where 256x256 central crop is used during testing.

\subsubsection{I3D on Charades}

This setup is used to show the advantages of MML regarding other approaches. Both D3D \cite{Stroud_2020_WACV} and MARS \cite{Crasto_2019_CVPR} deal with 3d-models, that is why we use the I3D ResNet-50 model \cite{carreira2017quo} to make a fair comparison with the mentioned methods.

Besides, Charades is the dataset with multiple corresponding classes per one clip, so we show how to extend the proposed MML to the multi-label task.

Optimizer, the number of epochs and other hyperparameters are taken from the standard config-file for the Charades training in \url{https://github.com/facebookresearch/SlowFast} without any changes. We use model trained on Kinetics-400 \cite{carreira2017quo} as a standard initialization instead of ImageNet initialization.

\begin{table*}[!t]
	\begin{minipage}{0.48\textwidth}
		\centering
		\caption{Same modality ML}
		\label{tab:sameml}
		\centering
		\begin{tabular}{|m{0.23\columnwidth}|c|m{0.23\columnwidth}|c|} 
			\hline \it First model & \it Top-1 / Top-5 & \it Second model & \it Top-1 / Top-5 \\ \hline
			\it RGB from ImageNet & 57.76 / 84.42 & \it RGB from ImageNet2 & 58.15 / 84.64 \\ \hline
			\it RGB from RGB & 57.84 / 84.55 & \it RGB from ImageNet & 60.20 / 86.33 \\ \hline
			\it RGB from RGB & \bf 60.54 / 86.23 & \it RGB from RGB2 & 60.47 / 86.08 \\ \thickhline
			\it Flow from ImageNet & 52.94 / 82.21 & \it Flow from ImageNet2 & 53.44 / 82.50 \\ \hline
			\it Flow from RGB & 57.58 / 85.17 & \it Flow from RGB2 & \bf 57.71 / 85.26 \\ \hline
		\end{tabular}
	\end{minipage}\hfill
	\begin{minipage}{0.48\textwidth}
		\centering
		\caption{I3D on Charades}
		\label{tab:i3d}
		\centering
		\begin{tabular}{|m{0.4\columnwidth}|c|c|} 
			\hline \it Training pipeline & \it RGB model mAP & \it Flow model mAP \\ \hline
			\it Ordinary training from Kinetics & 33.72 & 15.81 \\ \thickhline
			\it MARS training from Kinetics & 28.74 &  \\ \hline
			\it MARS training from RGB & 34.40 &  \\ \thickhline
			\it D3D training from Kinetics & 33.03 &  \\ \hline
			\it D3D training from RGB & 35.48 &  \\ \thickhline
			\it MML training from Kinetics & 33.84 & 17.34 \\ \hline
			\it MML training from RGB & \bf 35.96 & \bf 29.12 \\ \hline
		\end{tabular}
	\end{minipage}
\end{table*}

\subsection{Initialization}

An abbreviation "Flow from ImageNet"\ means that we initialize a model that takes Optical Flow as an input with the weights of the model trained on ImageNet. An abbreviation "Diff from RGB"\ means that we initialize a model that takes differences between RGB frames as an input with the weights of the model with RGB input trained on the current dataset using the cross-entropy loss and initialized by a model trained on ImageNet. We make other abbreviations in a similar way.

We do not include training from scratch into the ablation studies since this is a well-known fact that ImageNet initialization outperforms random initialization for the training of one-stream video models \cite{carreira2017quo,lin2019tsm}.

We can see from Table~\ref{tab:solo} that RGB initialization significantly outperforms ImageNet initialization in the case of ordinary cross-entropy training of the Flow and Diff models. At the same time, Flow initialization is useless for RGB models.

We apply MARS \cite{Crasto_2019_CVPR} and D3D \cite{Stroud_2020_WACV} approaches in both directions for RGB and Flow models. Table~\ref{tab:mars} shows that RGB initialization improves results in each scenario. It should be noted that both MARS and D3D approaches mainly target 3d-models. That is why the results of MARS training of "RGB from ImageNet"\ may be worse than the baseline ("RGB from ImageNet"\ using cross-entropy) since we use 2d-models.

Table~\ref{tab:rgbml} and~\ref{tab:flowml} are more representative. First, we train "RGB from ImageNet"\ and "Flow from ImageNet"\ models using cross-entropy. Then we train Flow and RGB models together using MML with all possible pairs of the initialization. The results of the RGB models trained using MML are presented in Table~\ref{tab:rgbml}. The results of the Flow models trained using MML are presented in Table~\ref{tab:flowml}.

As we can see, the middle values of each column in Table~\ref{tab:rgbml} are the best as well as the right values of each row in Table~\ref{tab:flowml}. Thus, the consistency of better initialization is preserved in the case of MML.

Finally, even if we train models on one modality using ML then RGB initialization is still the best. The first three rows and the next two rows of Table~\ref{tab:sameml} confirm that. We use abbreviations ImageNet2 and RGB2 to point out that we use different initialization obtained in the same way (KL loss is equal to zero otherwise).

\subsection{MML versus MARS and D3D} \label{subsec:versus}

Since MARS \cite{Crasto_2019_CVPR} and D3D \cite{Stroud_2020_WACV} works target mainly 3d-models, we use the I3D on Charades setup in this subsection.

It should be noted that ordinary KL loss implementation uses the "batchmean"\ regime of averaging, i.e.~we divide the sum of losses by the number of instances in one batch. However, we have to use the "mean"\ regime of averaging when we train the multi-label model using Binary Cross-Entropy losses (BCE), i.e.~we divide the sum of losses by the multiplication of two factors: the number of instances in one batch and the number of classes. See Appendix~\ref{app:c} as an explanation of this point.

By similar reasoning, we divide additional loss functions of MARS and D3D by the number of classes.

The mean Average Precision (mAP) results of all approaches are presented in Table~\ref{tab:i3d}. We can see again that RGB initialization improves the performance of each method. D3D is still better than MARS and MML is the best.

The right column in Table~\ref{tab:i3d} is empty for MARS and D3D approaches since these approaches do not modify the Optical Flow model during training.

We assume that performance correlates negatively with the strength of the supervision signal. Since we apply KL loss to probits, then any $l_2$ distance between logits is possible during MML. Thus, we weakly bound the feature extraction strategy of a network. In the case of D3D training, we minimize $l_2$ distance between logits only. Thus, D3D does not force a network to estimate the same features in contrast to MARS.

We want to stress that we can significantly improve a single-modality model different from RGB, e.g.~MML improves mAP of the Flow model by about 2 times. With some further research these findings may be very helpful for video recognition by event cameras \cite{gehrig2020video}.

\begin{table}[!t]	
	\centering
	\caption{MML and ML from ML}
	\label{tab:mml}
	\centering
	\begin{tabular}{|c|m{0.2\columnwidth}|c|m{0.2\columnwidth}|c|c|} 
		\hline & \it First model & \it Top-1 / Top-5 & \it Second model & \it Top-1 / Top-5 & \it Tag \\ \hline
		\it 1 & \it RGB from RGB & 60.82 / 86.75 & \it Flow from RGB & 57.95 / 85.44 & \bf A \\ \hline
		\it 2 & \it RGB from RGB & 60.88 / 86.86 & \it Flow from RGB2 & 57,87 / \textbf{85.53} & \\ \hline
		\it 3 & \it RGB from \bf A\it (RGB) & 61.18 / 86.81 & \it Flow from \bf A\it (RGB) & 58.02 / 85.49 & \bf B \\ \hline
		\it 4 & \it RGB from \bf B\it (RGB) & 61.15 / 86.81 & \it Flow from \bf B\it (RGB) & 57.96 / 85.30 &  \\ \hline
		\it 5 & \it RGB from RGB & 60.54 / 86.23 & \it RGB from RGB2 & 60.47 / 86.08 & \bf C \\ \hline
		\it 6 & \it RGB from \bf C\it (RGB) & 60.68 / 86.35 & \it RGB from \bf C\it (RGB2) & 60.88 / 86.44 &  \\ \hline
		\it 7 & \it RGB from \bf C\it (RGB) & \bf 61.30 / 86.99 & \it Flow from \bf C\it (RGB) & \textbf{58.36} / 85.49 &  \\ \hline
	\end{tabular}
\end{table}

\subsection{Relaunch of the ML and MML versus ML}

An abbreviation "RGB from \textbf{A}(RGB)"\ in Table~\ref{tab:mml} means that we initialize an RGB model with the weights of the RGB model that was trained by ML tagged as \textbf{A}.

Rows number 1 and number 3 from Table~\ref{tab:mml} demonstrate that relaunch of MML can improve the performance. At the same time, row number 4 demonstrates that the second relaunch is probably useless. Rows number 5 and number 6 demonstrate that the consistency is preserved for single-modality ML.

As we can see, the results of the RGB model trained using MML are better than the results of both RGB models trained using ML: row number 1 versus row number 5 from Table~\ref{tab:mml}. This consistency is preserved for the relaunch of ML: row number 6 versus row number 7.

Rows number 1 and number 2 demonstrate that initialization with different RGB weights does not significantly affect the performance of MML.

Finally, row number 7 compared to row number 3 demonstrates that it is better to use MML only as a second step. We believe that the reason for that the separation of advantages of ML itself and additional information from another modality. We extensively examine this effect in section~\ref{sec:ensemble}.

\begin{table*}[!t]
	\centering
	\caption{RGBDiff modality}
	\label{tab:diff}
	\centering
	\begin{tabular}{|c|c|c|c|c|c|c|} 
		\hline \it Row number & \it First model & \it Top-1 / Top-5 & \it Second model & \it Top-1 / Top-5 & \it Third model & \it Top-1 / Top-5 \\ \hline
		\it 1 & \it RGB from RGB & 60.54 / 86.23 & \it RGB from RGB2 & 60.47 / 86.08 & & \\ \hline
		\it 2 & \it RGB from RGB & 60.82 / \textbf{86.75} & \it Flow from RGB & 57.95 / 85.44 & & \\ \hline
		\it 3 & \it Flow from RGB & 57.58 / 85.17 & \it Flow from RGB & 57.71 / 85.26 & & \\ \hline
		\it 4 & \it Diff from RGB & 60.66 / 87.65 & \it Diff from RGB2 & 61.07 / 87.73 & & \\ \hline
		\it 5 & \it RGB from RGB & 60.52 / 86.52 & \it Diff from RGB & 62.13 / 87.57 & & \\ \hline
		\it 6 & \it RGB from RGB & \textbf{61.03} / 86.71 & \it Flow from RGB & \bf 58.03 / 85.61 & Diff from RGB & \bf 62.51 / 87.95 \\ \hline
		
	\end{tabular}
\end{table*}

\begin{figure*}[!t]
	\centering
	\includegraphics[width=\linewidth]{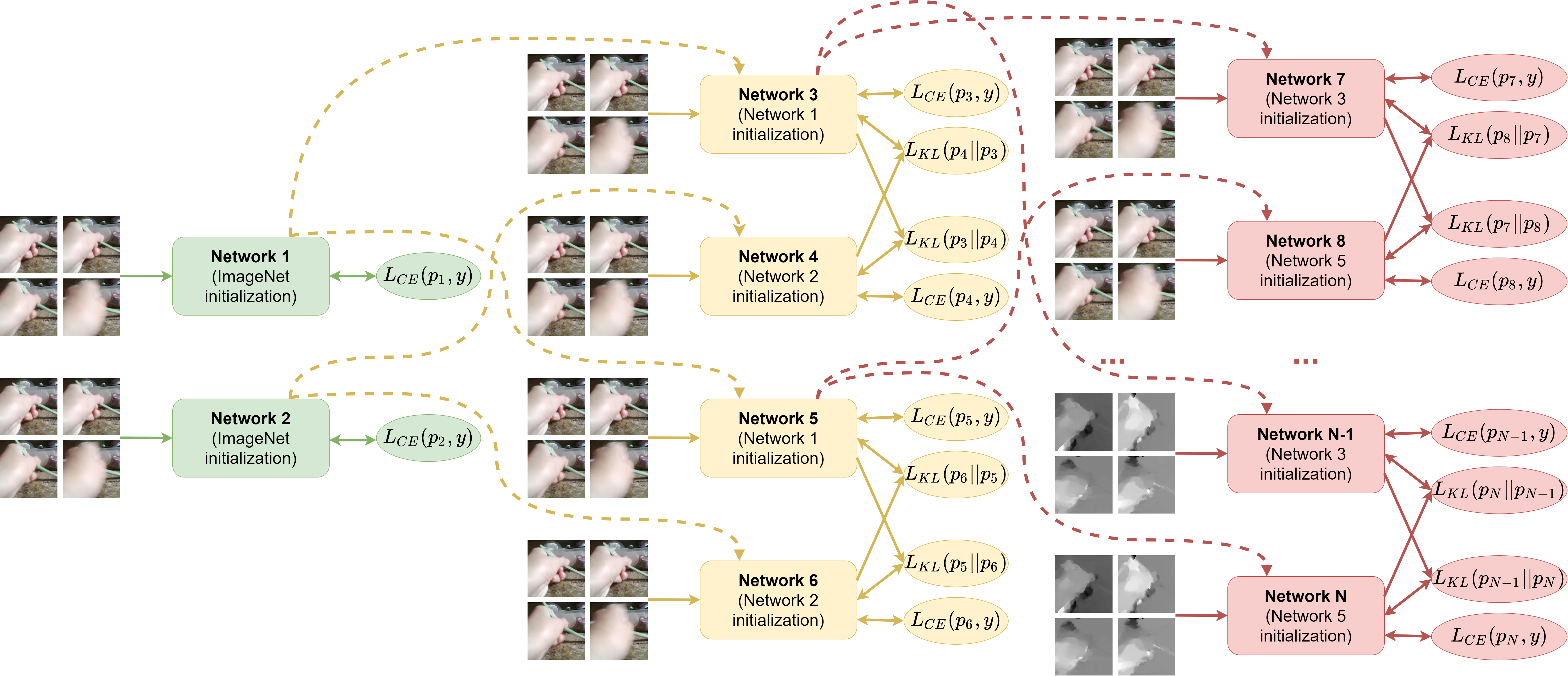}
	\caption{Best viewed in color. Solid arrows denote flows of data. Dashed arrows denote weights transferring for initialization.
		Green part: first, we train two networks with RGB input initialized by ImageNet weights using cross-entropy loss.
		Yellow part: next, we launch RGB-only Mutual Learning for two times. We use the weights from the first step as initialization for each launch of Mutual Learning. We have to use two launches for the second step because we need to obtain two models for which KL loss has not been optimized yet.
		Red part: finally, we apply single-modality Mutual Learning to each modality that we want to use in the ensemble. We use the weight from one model from each pair from the previous step as the initialization.}
	\label{fig:ensemble}
\end{figure*}

\subsection{Other modalities}

We expand our experiments to the Diff modality to examine the preservation of the consistency.

The last row from Table~\ref{tab:solo} confirms that RGB initialization is also useful for Diff model.

Row number 5 compared to rows number 4 and number 1 from Table~\ref{tab:diff} confirms that MML is not worse than or even better than single-modality ML in case of RGB and Diff modalities.

Finally, the comparison of row number 6 to rows 1--5 demonstrates that MML with all three modalities outperforms or is not worse than any other ML in terms of individual results for each modality.

\begin{table*}[!t]
	\centering
	\caption{SOTA on Something-Something-v2}
	\label{tab:sota}
	\centering
	\begin{tabular}{|c|c|c|m{0.07\linewidth}|m{0.1\linewidth}|m{0.07\linewidth}|m{0.07\linewidth}|m{0.07\linewidth}|m{0.07\linewidth}|} 
		\hline \it Solution & \it Ensemble & \it Base architecture & \it Number of input frames &\it Spatial crops $\times$ Temporal clips for prediction & \it Top-1 on validation & \it Top-5 on validation & \it Top-1 on test & \it Top-5 on test \\ \hline
		TSM \cite{lin2019tsm} & No & ResNet-50 & 8 & $1\times1$ & 59.1 & 85.6 & $-$ & $-$ \\ \hline
		TIN \cite{shao2020temporal} & No & ResNet-50 & 8 & $1\times1$ & 60.0 & 85.5 & $-$ & $-$ \\ \hline
		TPN \cite{yang2020temporal} & No & ResNet-50 & 8 & $1\times1$ & \bf 62.0 & $-$ & $-$ & $-$ \\ \hline
		\bf MML (ours) & No & ResNet-50 & 8 & $1\times1$ & 61.87 & \bf 87.32 & $-$ & $-$ \\ \thickhline
		STM \cite{jiang2019stm} & No & ResNet-50 & 8 & $3\times?$ & 62.3 & 88.8 & 61.3 & 88.4 \\ \thickhline
		W3 \cite{perez2020knowing} & No & ResNet-50 & 16 & $?\times2$ & \bf 66.5 & \bf 90.4 & $-$ & $-$ \\ \hline
		STM \cite{jiang2019stm} & No & ResNet-50 & 16 & $3\times?$ & 64.2 & 89.8 & 63.5 & 89.6 \\ \hline
		TPN \cite{yang2020temporal} & No & ResNet-101 & 16 & $3\times2$ & $-$ & $-$ & \bf 67.72 & 91.28 \\ \hline
		\bf MML (ours) & No & ResNet-101 & 16 & $1\times3$ & 65.9 & 90.15 & 66.83 & \bf 91.30 \\ \thickhline
		bLVNet-TAM RGB+Flow \cite{fan2019more} & Yes & ResNet-101 & 32+32 & $3\times10$ & 68.5 & 91.4 & 67.1 & 91.4 \\ \hline
		TSM RGB+Flow\cite{lin2019tsm} & Yes & ResNet-50 & 16+16 & $?\times?$ & 66.0 & 90.5 & 66.55 & 91.25 \\ \hline
		RGB-only ensemble (9702\_10347) by Anonymous & Yes & $-$ & $-$ & $?\times?$ & $-$ & $-$ & 68.18 & 91.26 \\ \hline
		TSM ResNet-101, RGB+Flow by Anonymous & Yes & ResNet-101 & $-$ & $?\times?$ & $-$ & $-$ & 67.71 & 91.95 \\ \hline
		\bf ML RGB+Flow (ours) & Yes & ResNet-101 & 16+16 & $1\times3$ & 68.16 & 91.69 & $-$ & $-$ \\ \hline
		\bf ML RGB+Flow+Diff (ours) & Yes & ResNet-101 & 16+16+16 & $1\times3$ & \bf 69.07 & \bf 92.07 & \bf 69.02 & \bf 92.70 \\ \hline
	\end{tabular}
\end{table*}

\section{Ensemble performance} \label{sec:ensemble}

The predictions of RGB and Flow models can be highly correlated since we train them using KL loss. Thus, an averaging of the predictions may perform worse than the averaging of ordinary RGB and Flow models trained using cross-entropy. The same logic is applicable to MARS or D3D training.

We show results of ensembles of two models in Appendix~\ref{app:b1} and some results of ensembles of three different models with RGB, Flow and Diff input modalities in Appendix~\ref{app:b2}.

The main conclusions are as follows: 
\begin{itemize}
	\item RGB models that do not use Optical Flow during training perform the best in ensemble with Flow models. Models trained using ML with RGB only are the first, RGB models trained using MML with RGBDiff are the second.
	
	\item RGB models that use Optical Flow during training are the worst in the ensemble with Flow models.
	
	Performance in the ensemble with Flow models from better to worse: MML, D3D, MARS. We believe that this order is caused by the same reasons that are mentioned in the subsection~\ref{subsec:versus}.

	\item The same behavior preserves when we combine Flow models with/without RGB signals in loss function during training with RGB models. The only point we want to stress is that "Flow from RGB"\ models still perform better than "Flow from ImageNet"\ models in ensembles with RGB models.
	
	\item It is also better to combine models trained using single-modality ML when we average the predictions of the RGB and Diff models. 
	
	\item An ensemble of RGB and Diff models can achieve results that are similar to the results of the RGB and Flow ensemble.
	
	\item Models trained using single-modality ML achieve the best results in the ensemble of three different modalities in our experiments. See Appendix~\ref{app:b2} for more details.	
	
\end{itemize}

Thus, although MML provides the best single-modality models, ordinary ML performs better for ensembles. Considering the aforementioned observations, we propose a pipeline for the best ensemble training that is depicted in Figure~\ref{fig:ensemble}.

First, we train two "RGB from ImageNet"\ models using cross-entropy. Second, we launch two single-modality ML procedures for the RGB models from the previous step. Finally, we train models using single-modality ML for each of three modalities (RGB, Flow, Diff) that we want to use in the ensemble. We use weights of the RGB models from the second step as an initialization for the third step. This is the reason why we have to launch two training procedures on the second step. KL loss is already optimized otherwise.

\section{Comparison to state-of-the-art}

Followed by the observations made above, we train TSM+ResNet-101 with 16 input frames per clip on Something-Something-v2 using the described pipelines. Thus, we obtain an enhanced RGB model trained by MML and three models with RGB, Flow and Diff inputs for the best ensemble according to section~\ref{sec:ensemble}. The results are available in Table~\ref{tab:sota}.

Our pipeline for the best ensemble achieves the SOTA results among the ones reported previously in the Something-Something-v2 benchmark.

We also make a comparison with other single-model solutions. There is only one single-model solution that outperforms our solution in one of two testing metrics in the Something-Something-v2 benchmark. This is a Temporal Pyramid Network \cite{yang2020temporal} that is several times heavier than TSM and uses more launches per one prediction.

For the simplest scenario, when we use ResNet-50 as a base architecture with 8 input frames and one launch per prediction, we achieve +2.77\% improvement of the top-1 performance without adding complexity for the inference.

We exclude STM \cite{jiang2019stm} model from the comparison since it uses the average of predictions for three spatial crops. That is a more accurate but also a more computationally expensive approach.

\section{Conclusion}

We present Mutual Modality Learning, the approach that enhances the performance of single-modality model by joint training with models based on other modalities. In addition, we show that the proper initialization of network weights boosts the performance of various training scenarios. We check that our proposal works for different models and datasets, even for multi-label tasks. Our experiments lead to state-of-the-art results in the Something-Something-v2 benchmark.

\bibliographystyle{plain}
\bibliography{root}

\begin{thebibliography}{10}

\bibitem{carreira2019short}
Joao Carreira, Eric Noland, Chloe Hillier, and Andrew Zisserman.
\newblock A short note on the kinetics-700 human action dataset.
\newblock {\em arXiv preprint arXiv:1907.06987}, 2019.

\bibitem{carreira2017quo}
Joao Carreira and Andrew Zisserman.
\newblock Quo vadis, action recognition? a new model and the kinetics dataset.
\newblock In {\em proceedings of the IEEE Conference on Computer Vision and
  Pattern Recognition}, pages 6299--6308, 2017.

\bibitem{Crasto_2019_CVPR}
Nieves Crasto, Philippe Weinzaepfel, Karteek Alahari, and Cordelia Schmid.
\newblock Mars: Motion-augmented rgb stream for action recognition.
\newblock In {\em The IEEE Conference on Computer Vision and Pattern
  Recognition (CVPR)}, June 2019.

\bibitem{deng2009imagenet}
Jia Deng, Wei Dong, Richard Socher, Li-Jia Li, Kai Li, and Li~Fei-Fei.
\newblock Imagenet: A large-scale hierarchical image database.
\newblock In {\em 2009 IEEE conference on computer vision and pattern
  recognition}, pages 248--255. Ieee, 2009.

\bibitem{fan2018end}
Lijie Fan, Wenbing Huang, Chuang Gan, Stefano Ermon, Boqing Gong, and Junzhou
  Huang.
\newblock End-to-end learning of motion representation for video understanding.
\newblock In {\em Proceedings of the IEEE Conference on Computer Vision and
  Pattern Recognition}, pages 6016--6025, 2018.

\bibitem{fan2019more}
Quanfu Fan, Chun-Fu~Richard Chen, Hilde Kuehne, Marco Pistoia, and David Cox.
\newblock More is less: Learning efficient video representations by big-little
  network and depthwise temporal aggregation.
\newblock In {\em Advances in Neural Information Processing Systems}, pages
  2264--2273, 2019.

\bibitem{feichtenhofer2019slowfast}
Christoph Feichtenhofer, Haoqi Fan, Jitendra Malik, and Kaiming He.
\newblock Slowfast networks for video recognition.
\newblock In {\em Proceedings of the IEEE International Conference on Computer
  Vision}, pages 6202--6211, 2019.

\bibitem{furlanello2018born}
Tommaso Furlanello, Zachary~C Lipton, Michael Tschannen, Laurent Itti, and
  Anima Anandkumar.
\newblock Born again neural networks.
\newblock {\em arXiv preprint arXiv:1805.04770}, 2018.

\bibitem{gehrig2020video}
Daniel Gehrig, Mathias Gehrig, Javier Hidalgo-Carri{\'o}, and Davide
  Scaramuzza.
\newblock Video to events: Recycling video datasets for event cameras.
\newblock In {\em Proceedings of the IEEE/CVF Conference on Computer Vision and
  Pattern Recognition}, pages 3586--3595, 2020.

\bibitem{goyal2017something}
Raghav Goyal, Samira~Ebrahimi Kahou, Vincent Michalski, Joanna Materzynska,
  Susanne Westphal, Heuna Kim, Valentin Haenel, Ingo Fruend, Peter Yianilos,
  Moritz Mueller-Freitag, et~al.
\newblock The "something something" video database for learning and evaluating
  visual common sense.
\newblock In {\em ICCV}, volume~1, page~5, 2017.

\bibitem{he2016deep}
Kaiming He, Xiangyu Zhang, Shaoqing Ren, and Jian Sun.
\newblock Deep residual learning for image recognition.
\newblock In {\em Proceedings of the IEEE conference on computer vision and
  pattern recognition}, pages 770--778, 2016.

\bibitem{hinton2015distilling}
Geoffrey Hinton, Oriol Vinyals, and Jeff Dean.
\newblock Distilling the knowledge in a neural network.
\newblock {\em arXiv preprint arXiv:1503.02531}, 2015.

\bibitem{jiang2019stm}
Boyuan Jiang, MengMeng Wang, Weihao Gan, Wei Wu, and Junjie Yan.
\newblock Stm: Spatiotemporal and motion encoding for action recognition.
\newblock In {\em Proceedings of the IEEE International Conference on Computer
  Vision}, pages 2000--2009, 2019.

\bibitem{kay2017kinetics}
Will Kay, Joao Carreira, Karen Simonyan, Brian Zhang, Chloe Hillier, Sudheendra
  Vijayanarasimhan, Fabio Viola, Tim Green, Trevor Back, Paul Natsev, et~al.
\newblock The kinetics human action video dataset.
\newblock {\em arXiv preprint arXiv:1705.06950}, 2017.

\bibitem{kuehne2011hmdb}
Hildegard Kuehne, Hueihan Jhuang, Est{\'\i}baliz Garrote, Tomaso Poggio, and
  Thomas Serre.
\newblock Hmdb: a large video database for human motion recognition.
\newblock In {\em 2011 International Conference on Computer Vision}, pages
  2556--2563. IEEE, 2011.

\bibitem{lin2019tsm}
Ji~Lin, Chuang Gan, and Song Han.
\newblock Tsm: Temporal shift module for efficient video understanding.
\newblock In {\em Proceedings of the IEEE International Conference on Computer
  Vision}, pages 7083--7093, 2019.

\bibitem{miech2019howto100m}
Antoine Miech, Dimitri Zhukov, Jean-Baptiste Alayrac, Makarand Tapaswi, Ivan
  Laptev, and Josef Sivic.
\newblock Howto100m: Learning a text-video embedding by watching hundred
  million narrated video clips.
\newblock In {\em Proceedings of the IEEE International Conference on Computer
  Vision}, pages 2630--2640, 2019.

\bibitem{perez2020knowing}
Juan-Manuel Perez-Rua, Brais Martinez, Xiatian Zhu, Antoine Toisoul, Victor
  Escorcia, and Tao Xiang.
\newblock Knowing what, where and when to look: Efficient video action modeling
  with attention.
\newblock {\em arXiv preprint arXiv:2004.01278}, 2020.

\bibitem{piergiovanni2019representation}
AJ~Piergiovanni and Michael~S Ryoo.
\newblock Representation flow for action recognition.
\newblock In {\em Proceedings of the IEEE Conference on Computer Vision and
  Pattern Recognition}, pages 9945--9953, 2019.

\bibitem{shao2020temporal}
Hao Shao, Shengju Qian, and Yu~Liu.
\newblock Temporal interlacing network.
\newblock {\em arXiv preprint arXiv:2001.06499}, 2020.

\bibitem{Sigurdsson2016HollywoodIH}
Gunnar~A. Sigurdsson, G{\"u}l Varol, Xiaolong Wang, Ali Farhadi, Ivan Laptev,
  and Abhinav Gupta.
\newblock Hollywood in homes: Crowdsourcing data collection for activity
  understanding.
\newblock In {\em ECCV}, 2016.

\bibitem{simonyan2014two}
Karen Simonyan and Andrew Zisserman.
\newblock Two-stream convolutional networks for action recognition in videos.
\newblock In {\em Advances in neural information processing systems}, pages
  568--576, 2014.

\bibitem{soomro2012ucf101}
Khurram Soomro, Amir~Roshan Zamir, and Mubarak Shah.
\newblock Ucf101: A dataset of 101 human actions classes from videos in the
  wild.
\newblock {\em arXiv preprint arXiv:1212.0402}, 2012.

\bibitem{Stroud_2020_WACV}
Jonathan Stroud, David Ross, Chen Sun, Jia Deng, and Rahul Sukthankar.
\newblock D3d: Distilled 3d networks for video action recognition.
\newblock In {\em The IEEE Winter Conference on Applications of Computer Vision
  (WACV)}, March 2020.

\bibitem{tran2015learning}
Du~Tran, Lubomir Bourdev, Rob Fergus, Lorenzo Torresani, and Manohar Paluri.
\newblock Learning spatiotemporal features with 3d convolutional networks.
\newblock In {\em Proceedings of the IEEE international conference on computer
  vision}, pages 4489--4497, 2015.

\bibitem{tran2018closer}
Du~Tran, Heng Wang, Lorenzo Torresani, Jamie Ray, Yann LeCun, and Manohar
  Paluri.
\newblock A closer look at spatiotemporal convolutions for action recognition.
\newblock In {\em Proceedings of the IEEE conference on Computer Vision and
  Pattern Recognition}, pages 6450--6459, 2018.

\bibitem{wang2016temporal}
Limin Wang, Yuanjun Xiong, Zhe Wang, Yu~Qiao, Dahua Lin, Xiaoou Tang, and Luc
  Van~Gool.
\newblock Temporal segment networks: Towards good practices for deep action
  recognition.
\newblock In {\em European conference on computer vision}, pages 20--36.
  Springer, 2016.

\bibitem{wang2019makes}
Weiyao Wang, Du~Tran, and Matt Feiszli.
\newblock What makes training multi-modal networks hard?
\newblock {\em arXiv preprint arXiv:1905.12681}, 2019.

\bibitem{wang2018non}
Xiaolong Wang, Ross Girshick, Abhinav Gupta, and Kaiming He.
\newblock Non-local neural networks.
\newblock In {\em Proceedings of the IEEE conference on computer vision and
  pattern recognition}, pages 7794--7803, 2018.

\bibitem{xie2018rethinking}
Saining Xie, Chen Sun, Jonathan Huang, Zhuowen Tu, and Kevin Murphy.
\newblock Rethinking spatiotemporal feature learning: Speed-accuracy trade-offs
  in video classification.
\newblock In {\em Proceedings of the European Conference on Computer Vision
  (ECCV)}, pages 305--321, 2018.

\bibitem{yang2020temporal}
Ceyuan Yang, Yinghao Xu, Jianping Shi, Bo~Dai, and Bolei Zhou.
\newblock Temporal pyramid network for action recognition.
\newblock {\em arXiv preprint arXiv:2004.03548}, 2020.

\bibitem{zach2007duality}
Christopher Zach, Thomas Pock, and Horst Bischof.
\newblock A duality based approach for realtime tv-l 1 optical flow.
\newblock In {\em Joint pattern recognition symposium}, pages 214--223.
  Springer, 2007.

\bibitem{zhang2016real}
Bowen Zhang, Limin Wang, Zhe Wang, Yu~Qiao, and Hanli Wang.
\newblock Real-time action recognition with enhanced motion vector cnns.
\newblock In {\em Proceedings of the IEEE conference on computer vision and
  pattern recognition}, pages 2718--2726, 2016.

\bibitem{zhang2018deep}
Ying Zhang, Tao Xiang, Timothy~M Hospedales, and Huchuan Lu.
\newblock Deep mutual learning.
\newblock In {\em Proceedings of the IEEE Conference on Computer Vision and
  Pattern Recognition}, pages 4320--4328, 2018.

\bibitem{zolfaghari2018eco}
Mohammadreza Zolfaghari, Kamaljeet Singh, and Thomas Brox.
\newblock Eco: Efficient convolutional network for online video understanding.
\newblock In {\em Proceedings of the European Conference on Computer Vision
  (ECCV)}, pages 695--712, 2018.

\end{thebibliography}

\begin{table*}[!h]
	\centering
	\caption{Testing with different sampling strategies}
	\label{tab:sampling}
	\centering
	\begin{tabular}{|p{0.1\linewidth}|p{0.1\linewidth}|p{0.1\linewidth}|p{0.1\linewidth}|p{0.1\linewidth}|p{0.1\linewidth}|p{0.1\linewidth}|p{0.1\linewidth}|p{0.1\linewidth}|} 
		\hline \it Sampling during training & \it Dense 0 + Uniform 1 & \it Dense 0 + Uniform 2 & \it Dense 1 + Uniform 0 & \it Dense 2 + Uniform 0 & \it Dense 1 + Uniform 1 & \it Dense 2 + Uniform 1 & \it Dense 1 + Uniform 2 & \it Dense 2 + Uniform 2 \\ \hline
		\it Dense sampling & 57.33 / 84.51 & 58.79 / 85.68 & 57.61 / 84.98 & \bf 58.70 / 85.66 & 59.71 / 86.57 & 60.07 / 86.47 & 60.11 / 86.56 & 60.27 / 86.61 \\ \hline 
		\it Uniform sampling & 59.86 / \textbf{86.14} & 61.16 / \textbf{87.03} & 56.25 / 83.72 & 56.20 / 84.18 & 60.64 / 85.58 & 59.69 / 86.38 & 61.50 / 87.32 & 61.03 / \textbf{87.10} \\ \hline 
		\it Both samplings & \textbf{60.11} / 85.79 & \textbf{61.38} / 86.82 & \bf 57.80 / 85.05 & 58.66 / 85.32 & \bf 61.10 / 86.66 & \bf 61.01 / 86.57 & \bf \underline{61.71 / 87.40} & \textbf{61.59} / 86.97 \\ \hline 	
	\end{tabular}
\end{table*}

\newpage

\appendices

\section{Sampling strategy} \label{app:a}

The common procedure for the Something-Something-v2 final testing is an averaging of two predictions for each video. For each prediction, we use central full-resolution crop and uniform sampling: we use frames with numbers $\left\{\left\lfloor\frac{0\cdot T}{N}\right\rfloor,\ldots,\left\lfloor\frac{(N-1)\cdot T}{N}\right\rfloor\right\}$ for the first prediction and frames with numbers $\left\{\left\lfloor\frac{0.5\cdot T}{N}\right\rfloor,\ldots,\left\lfloor\frac{(N-0.5)\cdot T}{N}\right\rfloor\right\}$ for the second prediction, where $T$ is a total number of frames for current modality and $N$ is the shape of the temporal dimension of the input. Note that uniform sampling, unlike dense sampling, allows any period between input frames and depends on the total length of the video.

We found out that the use of more than two temporal crops with the same sampling strategy or more number of spatial crops insignificantly improves the validation results. At the same time, the use of different sampling strategies during testing significantly improves results regardless of the sampling strategy during training. That is why we incorporate both samplings into training. The median testing results for full-resolution central crops testing are shown in Table~\ref{tab:sampling}.
Label "Dense $k$ + Uniform $m$"\ means that we use $k+m$ predictions per video using frames with numbers $\left\{\left\lfloor\frac{i\cdot T'}{k}\right\rfloor,\left\lfloor\frac{i\cdot T'}{k}+\tau\right\rfloor,\ldots,\left\lfloor\frac{i\cdot T'}{k}+\tau\cdot(N-1)\right\rfloor\right\},\ i\in\{0,\ldots,k-1\}$ when $k>1$ or frames with numbers $\left\{\left\lfloor\frac{ T'}{2}\right\rfloor,\left\lfloor\frac{ T'}{2}+\tau\right\rfloor,\ldots,\left\lfloor\frac{ T'}{2}+\tau\cdot(N-1)\right\rfloor\right\}$ when $k=1$ and frames with numbers $\left\{\left\lfloor\frac{i/m\cdot T}{N}\right\rfloor,\ldots,\left\lfloor\frac{(N-1+i/m)\cdot T}{N}\right\rfloor\right\},\ i\in\{0,\ldots,m-1\}$. Here $T$ is a total number of frames for current modality, $N$ is the shape of the temporal dimension of the input, $\tau < \frac{T}{N-1}$ is a dense for the dense sampling and $T'=T-\tau\cdot(N-1)$. Note that there is no random nature in frame numbers during testing.

We make the next conclusions based on Table~\ref{tab:sampling}:
\begin{itemize}
	\item Dense sampling training is not suitable for the Something-Something-v2.
	\item Uniform sampling training and Both samplings training are nearly equal if we use prediction for Uniform sampling.
	\item Both samplings training outperforms Uniform sampling strategy by up to one percent when tested with both strategies.
	\item It is better to average predictions for two Uniform samplings and one Dense sampling during testing.
\end{itemize}

\section{Loss modification for the BCE training} \label{app:c}

The ordinary implementation of the KL loss divides the sum of $B\cdot N$ terms by the $B$, where $B$ is a batch size and $N$ is a number of classes. The reason for that is that ordinary Cross-Entropy loss also divides the sum of $B\cdot N$ terms by the $B$, which can be unobvious:

$$L_{CE}=\frac{1}{B}\cdot\sum_{b=1}^{B} -\log{\frac{e^{l_b^{\text{gt}_b}}}{\sum_{j=1}^{N}e^{l_b^j}}}=$$
\begin{equation}=\frac{1}{B}\cdot\sum_{b=1}^{B}\sum_{i=1}^{N}-y_b^i\cdot\log{\frac{e^{l_b^{i}}}{\sum_{j=1}^{N}e^{l_b^j}}}=\end{equation}
$$=\frac{1}{B}\cdot\sum_{b=1}^{B}\sum_{i=1}^{N}-y_b^i\cdot\log{p_b^i}=\frac{1}{B}\cdot H(y,p).$$

Here $l_b^i$ is a predicted logit for the class number $i$ for the instance number $b$, $\text{gt}_b$ --- ground truth class for the instance number $b$, $y_b^i=I_{\text{gt}_b}(i)$ and $p_b^i=\frac{e^{l_b^{i}}}{\sum_{j=1}^{N}e^{l_b^j}}$.

So the magnitudes of the CE loss and KL loss are the same. Since the multi-label BCE loss is divided by the $B\cdot N$:
\begin{equation}L_{mlBCE} =\end{equation}
$$\frac{1}{B\cdot N}\cdot\sum_{b=1}^{B}\sum_{i=1}^{N}-\left(y_b^i\cdot\log{\sigma(l_b^i)}+(1-y_b^i)\cdot\log{\left(1-\sigma(l_b^i)\right)}\right),$$
then we divide the KL loss by the $B\cdot N$ to make the magnitudes the same again.

The authors of the MARS and D3D approaches found the best weights for their loss functions in the case of the Cross-Entropy training (50 for MARS and 1 for D3D).  Our experiments confirm that additional division of the loss by the number of classes improves the performance of these two methods in the case of multi-label training according to the reasoning made above.

\section{Ensembles}

\subsection{Ensembles of two models} \label{app:b1}

\begin{table*}[!h]
	\centering
	\caption{Ensemble of RGB and Flow models}
	\label{tab:rgbflow}
	\centering
	\begin{tabular}{|p{0.06\textwidth}|p{0.06\textwidth}|p{0.06\textwidth}|p{0.06\textwidth}|p{0.06\textwidth}|p{0.06\textwidth}|p{0.06\textwidth}|p{0.06\textwidth}|p{0.06\textwidth}|p{0.06\textwidth}|p{0.06\textwidth}|p{0.06\textwidth}|p{0.06\textwidth}|p{0.06\textwidth}|}
		\hline \diagbox[innerwidth=0.06\textwidth]{Flow}{RGB} & \it CE from ImageNet & \it MARS from RGB & \it D3D from RGB & \it MML with Flow from RGB (\textbf{A}) & \it MML from RGB with Flow from Flow & \it ML from RGB 1 (\textbf{B}) & \it ML from RGB 2 & \it ML from \textbf{B} 1 & \it ML from \textbf{B} 2 & \it MML fwith Flow from \textbf{A} & \it MML with Flow from \textbf{B} & \it MML with Diff from RGB & \it MML with Flow and Diff from RGB \\ \hline
		\it CE from ImageNet & \cellcolor[RGB]{255,255,255} 63.67 / 88.41 & \cellcolor[RGB]{91,155,213} 60.40 / 86.68 & \cellcolor[RGB]{157,195,230} 61.72 / 87.15 & \cellcolor[RGB]{204,224,242} 62.65 / 88.03 & \cellcolor[RGB]{199,221,241} 62.56 / 88.02 & \cellcolor[RGB]{255,149,149} 64.62 / 89.14 & \cellcolor[RGB]{255,181,181} 64.33 / 88.76 & \cellcolor[RGB]{255,168,168} 64.45 / 88.82 & \cellcolor[RGB]{255,171,171} 64.42 / 88.97 & \cellcolor[RGB]{206,225,243} 62.70 / 88.09 & \cellcolor[RGB]{219,233,246} 62.96 / 88.19 & \cellcolor[RGB]{255,244,244} 63.77 / 88.64 & \cellcolor[RGB]{237,244,250} 63.31 / 88.29 \\ \hline 
		\it CE from RGB & \cellcolor[RGB]{255,247,247} 63.74 / 88.73 & \cellcolor[RGB]{130,179,223} 61.18 / 87.42 & \cellcolor[RGB]{177,208,235} 62.12 / 87.91 & \cellcolor[RGB]{216,231,245} 62.89 / 88.59 & \cellcolor[RGB]{211,228,244} 62.79 / 88.41 & \cellcolor[RGB]{255,180,180} 64.34 / 89.44 & \cellcolor[RGB]{255,175,175} 64.39 / 89.21 & \cellcolor[RGB]{255,160,160} 64.52 / 89.45 & \cellcolor[RGB]{255,168,168} 64.45 / 89.35 & \cellcolor[RGB]{218,233,246} 62.94 / 88.68 & \cellcolor[RGB]{228,239,248} 63.14 / 88.68 & \cellcolor[RGB]{255,218,218} 64.00 / 88.94 & \cellcolor[RGB]{230,240,249} 63.18 / 88.73 \\ \hline 
		\it MARS from RGB & \cellcolor[RGB]{184,212,237} 62.26 / 87.81 & \cellcolor[RGB]{152,192,229} 61.61 / 87.24 & \cellcolor[RGB]{167,201,233} 61.92 / 87.66 & \cellcolor[RGB]{190,215,238} 62.37 / 88.25 & \cellcolor[RGB]{196,219,240} 62.50 / 88.19 & \cellcolor[RGB]{250,252,254} 63.57 / 88.84 & \cellcolor[RGB]{242,247,252} 63.42 / 88.57 & \cellcolor[RGB]{252,253,254} 63.62 / 88.75 & \cellcolor[RGB]{250,252,254} 63.58 / 88.71 & \cellcolor[RGB]{210,228,244} 62.78 / 88.42 & \cellcolor[RGB]{215,231,245} 62.88 / 88.49 & \cellcolor[RGB]{220,234,246} 62.98 / 88.44 & \cellcolor[RGB]{202,223,242} 62.62 / 88.37 \\ \hline 
		\it MARS from ImageNet & \cellcolor[RGB]{200,221,241} 62.57 / 87.67 & \cellcolor[RGB]{135,182,224} 61.28 / 87.21 & \cellcolor[RGB]{167,201,233} 61.92 / 87.71 & \cellcolor[RGB]{199,221,241} 62.55 / 88.08 & \cellcolor[RGB]{202,223,241} 62.61 / 88.02 & \cellcolor[RGB]{255,248,248} 63.73 / 88.80 & \cellcolor[RGB]{252,253,254} 63.61 / 88.62 & \cellcolor[RGB]{255,245,245} 63.76 / 88.76 & \cellcolor[RGB]{255,252,252} 63.70 / 88.81 & \cellcolor[RGB]{217,232,245} 62.92 / 88.26 & \cellcolor[RGB]{219,233,246} 62.96 / 88.38 & \cellcolor[RGB]{235,243,250} 63.28 / 88.47 & \cellcolor[RGB]{207,226,243} 62.72 / 88.31 \\ \hline 
		\it ML with Flow from ImageNet & \cellcolor[RGB]{242,247,252} 63.41 / 88.44 & \cellcolor[RGB]{108,165,217} 60.73 / 86.72 & \cellcolor[RGB]{183,211,237} 62.24 / 87.35 & \cellcolor[RGB]{228,238,248} 63.13 / 88.26 & \cellcolor[RGB]{223,235,247} 63.03 / 88.25 & \cellcolor[RGB]{255,120,120} 64.88 / 89.23 & \cellcolor[RGB]{255,144,144} 64.67 / 88.93 & \cellcolor[RGB]{255,147,147} 64.64 / 89.13 & \cellcolor[RGB]{255,155,155} 64.57 / 89.33 & \cellcolor[RGB]{237,244,251} 63.32 / 88.43 & \cellcolor[RGB]{253,254,255} 63.64 / 88.46 & \cellcolor[RGB]{255,184,184} 64.31 / 88.79 & \cellcolor[RGB]{249,252,254} 63.56 / 88.38 \\ \hline 
		\it ML Flow from RGB 1 & \cellcolor[RGB]{255,222,222} 63.97 / 88.93 & \cellcolor[RGB]{167,201,232} 61.91 / 87.78 & \cellcolor[RGB]{210,228,244} 62.78 / 88.31 & \cellcolor[RGB]{255,247,247} 63.74 / 89.01 & \cellcolor[RGB]{251,253,254} 63.60 / 88.91 & \cellcolor[RGB]{255,86,86} 65.19 / 89.77 & \cellcolor[RGB]{255,109,109} 64.98 / 89.45 & \cellcolor[RGB]{255,84,84} 65.20 / 89.74 & \cellcolor[RGB]{255,100,100} 65.06 / 89.72 & \cellcolor[RGB]{255,247,247} 63.74 / 89.09 & \cellcolor[RGB]{255,209,209} 64.08 / 89.20 & \cellcolor[RGB]{255,145,145} 64.66 / 89.22 & \cellcolor[RGB]{255,200,200} 64.16 / 89.16 \\ \hline 
		\it ML Flow from RGB 2 & \cellcolor[RGB]{255,193,193} 64.23 / 88.99 & \cellcolor[RGB]{173,205,234} 62.04 / 87.82 & \cellcolor[RGB]{218,232,245} 62.93 / 88.31 & \cellcolor[RGB]{255,230,230} 63.89 / 88.93 & \cellcolor[RGB]{254,255,255} 63.66 / 88.94 & \cellcolor[RGB]{255,80,80} 65.24 / 89.72 & \cellcolor[RGB]{255,94,94} 65.11 / 89.48 & \cellcolor[RGB]{255,98,98} 65.08 / 89.78 & \cellcolor[RGB]{255,97,97} 65.09 / 89.77 & \cellcolor[RGB]{255,216,216} 64.02 / 89.19 & \cellcolor[RGB]{255,195,195} 64.21 / 89.16 & \cellcolor[RGB]{255,126,126} 64.83 / 89.41 & \cellcolor[RGB]{255,187,187} 64.28 / 89.09 \\ \hline 
		\it MML with RGB from RGB & \cellcolor[RGB]{245,249,252} 63.47 / 88.41 & \cellcolor[RGB]{164,200,232} 61.86 / 87.62 & \cellcolor[RGB]{202,223,242} 62.62 / 88.03 & \cellcolor[RGB]{232,241,249} 63.22 / 88.53 & \cellcolor[RGB]{240,246,251} 63.38 / 88.55 & \cellcolor[RGB]{255,130,130} 64.79 / 89.35 & \cellcolor[RGB]{255,168,168} 64.45 / 88.91 & \cellcolor[RGB]{255,135,135} 64.75 / 89.32 & \cellcolor[RGB]{255,157,157} 64.55 / 89.21 & \cellcolor[RGB]{248,251,253} 63.53 / 88.75 & \cellcolor[RGB]{255,228,228} 63.91 / 88.72 & \cellcolor[RGB]{255,183,183} 64.32 / 88.81 & \cellcolor[RGB]{255,238,238} 63.82 / 88.78 \\ \hline 
		\it MML with RGB from \textbf{A} & \cellcolor[RGB]{255,243,243} 63.78 / 88.65 & \cellcolor[RGB]{175,206,235} 62.08 / 87.69 & \cellcolor[RGB]{216,231,245} 62.89 / 88.21 & \cellcolor[RGB]{250,252,254} 63.57 / 88.83 & \cellcolor[RGB]{251,253,254} 63.60 / 88.71 & \cellcolor[RGB]{255,109,109} 64.98 / 89.47 & \cellcolor[RGB]{255,139,139} 64.71 / 89.24 & \cellcolor[RGB]{255,121,121} 64.87 / 89.48 & \cellcolor[RGB]{255,111,111} 64.96 / 89.40 & \cellcolor[RGB]{250,252,254} 63.57 / 88.88 & \cellcolor[RGB]{255,214,214} 64.04 / 88.91 & \cellcolor[RGB]{255,179,179} 64.35 / 89.16 & \cellcolor[RGB]{255,235,235} 63.85 / 88.99 \\ \hline 
		\it MML with RGB from \textbf{B} & \cellcolor[RGB]{255,239,239} 63.81 / 88.73 & \cellcolor[RGB]{181,210,236} 62.20 / 87.71 & \cellcolor[RGB]{219,233,246} 62.95 / 88.04 & \cellcolor[RGB]{255,252,252} 63.70 / 88.76 & \cellcolor[RGB]{255,247,247} 63.74 / 88.70 & \cellcolor[RGB]{255,118,118} 64.90 / 89.37 & \cellcolor[RGB]{255,130,130} 64.79 / 89.20 & \cellcolor[RGB]{255,120,120} 64.88 / 89.27 & \cellcolor[RGB]{255,128,128} 64.81 / 89.40 & \cellcolor[RGB]{255,220,220} 63.98 / 88.78 & \cellcolor[RGB]{255,218,218} 64.00 / 88.84 & \cellcolor[RGB]{255,187,187} 64.28 / 88.95 & \cellcolor[RGB]{255,209,209} 64.08 / 88.97 \\ \hline 
		\it MML with RGB and Diff from RGB & \cellcolor[RGB]{252,253,254} 63.62 / 88.61 & \cellcolor[RGB]{168,202,233} 61.94 / 87.56 & \cellcolor[RGB]{200,221,241} 62.57 / 88.14 & \cellcolor[RGB]{242,247,252} 63.41 / 88.64 & \cellcolor[RGB]{246,250,253} 63.50 / 88.64 & \cellcolor[RGB]{255,118,118} 64.90 / 89.55 & \cellcolor[RGB]{255,139,139} 64.71 / 89.20 & \cellcolor[RGB]{255,134,134} 64.76 / 89.42 & \cellcolor[RGB]{255,162,162} 64.50 / 89.48 & \cellcolor[RGB]{251,253,254} 63.59 / 88.81 & \cellcolor[RGB]{255,238,238} 63.82 / 88.89 & \cellcolor[RGB]{255,170,170} 64.43 / 89.06 & \cellcolor[RGB]{253,254,255} 63.64 / 88.83 \\ \hline 
	\end{tabular}
\end{table*}

\begin{table*}[!h]
	\centering
	\caption{Ensemble of RGB and Diff models}
	\label{tab:rgbdiff}
	\centering
	\begin{tabular}{|p{0.06\textwidth}|p{0.06\textwidth}|p{0.06\textwidth}|p{0.06\textwidth}|p{0.06\textwidth}|p{0.06\textwidth}|p{0.06\textwidth}|p{0.06\textwidth}|p{0.06\textwidth}|p{0.06\textwidth}|p{0.06\textwidth}|p{0.06\textwidth}|p{0.06\textwidth}|p{0.06\textwidth}|}
		\hline \diagbox[innerwidth=0.06\textwidth]{Diff}{RGB} & \it CE from ImageNet & \it MARS from RGB & \it D3D from RGB & \it MML with Flow from RGB (\textbf{A}) & \it MML from RGB with Flow from Flow & \it ML from RGB 1 (\textbf{B}) & \it ML from RGB 2 & \it ML from \textbf{B} 1 & \it ML from \textbf{B} 2 & \it MML fwith Flow from \textbf{A} & \it MML with Flow from \textbf{B} & \it MML with Diff from RGB & \it MML with Flow and Diff from RGB \\ \hline
		\it CE from ImageNet & \cellcolor[RGB]{182,210,236} 63.26 / 88.46 & \cellcolor[RGB]{178,208,235} 63.24 / 88.29 & \cellcolor[RGB]{234,242,250} 63.55 / 88.59 & \cellcolor[RGB]{255,218,218} 63.97 / 88.89 & \cellcolor[RGB]{255,226,226} 63.90 / 88.77 & \cellcolor[RGB]{255,147,147} 64.54 / 89.33 & \cellcolor[RGB]{255,182,182} 64.26 / 89.01 & \cellcolor[RGB]{255,150,150} 64.52 / 89.27 & \cellcolor[RGB]{255,154,154} 64.48 / 89.14 & \cellcolor[RGB]{255,195,195} 64.15 / 89.03 & \cellcolor[RGB]{255,210,210} 64.03 / 89.07 & \cellcolor[RGB]{255,221,221} 63.94 / 88.86 & \cellcolor[RGB]{255,220,220} 63.95 / 88.98 \\ \hline 
		\it CE from RGB & \cellcolor[RGB]{153,193,229} 63.10 / 88.37 & \cellcolor[RGB]{178,208,235} 63.24 / 88.44 & \cellcolor[RGB]{237,244,250} 63.57 / 88.66 & \cellcolor[RGB]{250,252,254} 63.64 / 88.61 & \cellcolor[RGB]{255,231,231} 63.86 / 88.59 & \cellcolor[RGB]{255,168,168} 64.37 / 88.91 & \cellcolor[RGB]{255,210,210} 64.03 / 88.70 & \cellcolor[RGB]{255,146,146} 64.55 / 88.95 & \cellcolor[RGB]{255,194,194} 64.16 / 88.90 & \cellcolor[RGB]{255,230,230} 63.87 / 88.91 & \cellcolor[RGB]{255,214,214} 64.00 / 88.80 & \cellcolor[RGB]{255,233,233} 63.85 / 88.55 & \cellcolor[RGB]{255,235,235} 63.83 / 88.59 \\ \hline 
		\it ML from RGB 1 & \cellcolor[RGB]{255,254,254} 63.68 / 88.45 & \cellcolor[RGB]{255,198,198} 64.13 / 88.83 & \cellcolor[RGB]{255,200,200} 64.11 / 88.88 & \cellcolor[RGB]{255,153,153} 64.49 / 89.13 & \cellcolor[RGB]{255,150,150} 64.52 / 89.06 & \cellcolor[RGB]{255,106,106} 64.87 / 89.38 & \cellcolor[RGB]{255,151,151} 64.51 / 88.96 & \cellcolor[RGB]{255,84,84} 65.05 / 89.22 & \cellcolor[RGB]{255,107,107} 64.86 / 89.22 & \cellcolor[RGB]{255,136,136} 64.63 / 89.19 & \cellcolor[RGB]{255,122,122} 64.74 / 89.32 & \cellcolor[RGB]{255,177,177} 64.30 / 88.92 & \cellcolor[RGB]{255,163,163} 64.41 / 88.95 \\ \hline 
		\it ML from RGB 2 & \cellcolor[RGB]{255,230,230} 63.87 / 88.59 & \cellcolor[RGB]{255,178,178} 64.29 / 89.03 & \cellcolor[RGB]{255,173,173} 64.33 / 89.05 & \cellcolor[RGB]{255,145,145} 64.56 / 89.33 & \cellcolor[RGB]{255,121,121} 64.75 / 89.25 & \cellcolor[RGB]{255,99,99} 64.93 / 89.33 & \cellcolor[RGB]{255,107,107} 64.86 / 89.04 & \cellcolor[RGB]{255,80,80} 65.08 / 89.26 & \cellcolor[RGB]{255,106,106} 64.87 / 89.36 & \cellcolor[RGB]{255,87,87} 65.02 / 89.43 & \cellcolor[RGB]{255,100,100} 64.92 / 89.36 & \cellcolor[RGB]{255,135,135} 64.64 / 89.13 & \cellcolor[RGB]{255,118,118} 64.77 / 89.29 \\ \hline 
		\it MML with RGB from RGB & \cellcolor[RGB]{91,155,213} 62.75 / 88.12 & \cellcolor[RGB]{255,251,251} 63.70 / 88.62 & \cellcolor[RGB]{255,248,248} 63.73 / 88.66 & \cellcolor[RGB]{241,246,251} 63.59 / 88.71 & \cellcolor[RGB]{255,224,224} 63.92 / 88.73 & \cellcolor[RGB]{255,171,171} 64.35 / 89.07 & \cellcolor[RGB]{255,253,253} 63.69 / 88.61 & \cellcolor[RGB]{255,177,177} 64.30 / 88.89 & \cellcolor[RGB]{255,217,217} 63.98 / 88.90 & \cellcolor[RGB]{255,198,198} 64.13 / 88.83 & \cellcolor[RGB]{255,209,209} 64.04 / 89.00 & \cellcolor[RGB]{232,241,249} 63.54 / 88.47 & \cellcolor[RGB]{255,249,249} 63.72 / 88.64 \\ \hline 
		\it MML with RGB and Flow from RGB & \cellcolor[RGB]{221,234,246} 63.48 / 88.40 & \cellcolor[RGB]{203,223,242} 63.38 / 88.40 & \cellcolor[RGB]{244,248,252} 63.61 / 88.61 & \cellcolor[RGB]{255,198,198} 64.13 / 88.97 & \cellcolor[RGB]{255,195,195} 64.15 / 88.91 & \cellcolor[RGB]{255,104,104} 64.89 / 89.42 & \cellcolor[RGB]{255,157,157} 64.46 / 88.99 & \cellcolor[RGB]{255,101,101} 64.91 / 89.35 & \cellcolor[RGB]{255,150,150} 64.52 / 89.32 & \cellcolor[RGB]{255,177,177} 64.30 / 89.12 & \cellcolor[RGB]{255,154,154} 64.48 / 89.20 & \cellcolor[RGB]{255,192,192} 64.18 / 88.91 & \cellcolor[RGB]{255,192,192} 64.18 / 89.01 \\ \hline 
	\end{tabular}
\end{table*}

\begin{table*}[!h]
	\centering
	\caption{The relevance for the ensemble with other modalities}
	\label{tab:sums}
	\centering
	\begin{tabular}{|p{0.2\linewidth}|c"p{0.2\linewidth}|c"p{0.2\linewidth}|c|} 
		\hline RGB model & Sum of positions & Flow model & Sum of position & Diff model & Sum of positions \\ \hline
		ML from \textbf{B} 1 & 9486 & ML Flow from RGB 2 & 22434 & ML from RGB 2 & 42079 \\ \hline
		ML from RGB 1 (\textbf{B}) & 10711 & ML Flow from RGB 1 & 25793 & ML from RGB 1 & 49533 \\ \hline
		ML from \textbf{B} 2 & 13112 & MML with RGB from \textbf{A} & 28193 & CE from RGB & 67475 \\ \hline
		ML from RGB 2 & 15670 & MML with RGB from \textbf{B} & 28954 & MML with RGB from RGB & 67528 \\ \hline
		MML with Diff from RGB & 26918 & ML with Flow from ImageNet & 28972 & MML with RGB and Flow from RGB & 69895 \\ \hline
		MML with Flow from \textbf{B} & 28069 & CE from ImageNet & 29238 & CE from ImageNet & 71179 \\ \hline
		MML with Flow and Diff from RGB & 31281 & CE from RGB & 32372 &  &  \\ \hline
		CE from ImageNet & 31793 & MML with RGB and Diff from RGB & 32659 &  &  \\ \hline
		MML fwith Flow from \textbf{A} & 31968 & MML with RGB from RGB & 34290 &  &  \\ \hline
		MML from RGB with Flow from Flow & 35475 & MARS from ImageNet & 50577 &  &  \\ \hline
		MML with Flow from RGB (\textbf{A}) & 36167 & MARS from RGB & 54171 &  &  \\ \hline
		D3D from RGB  & 46044 &  &  &  &  \\ \hline
		MARS from RGB  & 50959 &  &  &  &  \\ \hline
	\end{tabular}
\end{table*}

Results of the ensembles of RGB and Flow models are depicted in Table~\ref{tab:rgbflow}. Results of the ensembles of RGB and Diff models are depicted in Table~\ref{tab:rgbdiff}.

"MML with Flow from RGB (\textbf{A})"\ in the first row means that we use the model with RGB input that was jointly trained using Mutual Modality Learning with the model with Optical Flow input using RGB initialization for both models. Tag \textbf{A} means that we use the weights of this model as initialization for other models in the table.

"ML from \textbf{B} 2"\ in the first row means that we use the second model with RGB input that was jointly trained using Mutual Learning with the other (the first) model with RGB input. Both models were initialized by the weights obtained by the procedure with tag \textbf{B}

We color the cell on the intersection of the column and the row that are marked "CE from ImageNet"\ in Table~\ref{tab:rgbflow} as white since it is the baseline ensemble. The more intense red color is, the higher the top-1 value for the ensemble is. The more intense light blue color is, the lower the top-1 value for the ensemble is.

The analysis of the tables is in section~\ref{sec:ensemble}.

\subsection{Ensembles of three models} \label{app:b2}

We evaluate the validation results for each combination of three models with different input modalities. We sort all the results of the ensembles of three models by the descending order. We show the sum of all indexes of positions for each model in Table~\ref{tab:sums}. So, the smaller value stands in the table the better model is in ensemble with two other modalities. Note that the magnitude of sums vary across the input modalities since there are different numbers of models for each modality are tested.

\end{document}